\documentclass[10pt,twocolumn,letterpaper]{article}

\usepackage{ijcb}
\usepackage[accsupp]{axessibility} 
\usepackage{times}
\usepackage{epsfig}
\usepackage{graphicx}
\usepackage{amsmath}
\usepackage{amssymb}

\usepackage[skip=0.5\baselineskip]{caption}
\usepackage{ijcb}
\usepackage{times}
\usepackage{epsfig}
\usepackage{graphicx}
\usepackage{amsmath}
\usepackage{amssymb}
\usepackage{mathtools, nccmath}
\usepackage[table,xcdraw]{xcolor}

\usepackage{multirow}

\ijcbfinalcopy 

\ifijcbfinal\fi
\begin{document}

\title{Deep Boosting Multi-Modal Ensemble Face Recognition with Sample-Level Weighting}

\author{ Sahar Rahimi Malakshan, Mohammad Saeed Ebrahimi Saadabadi,\\
Nima Najafzadeh, and Nasser M. Nasrabadi\\
{\tt\small{ sr00033, me00018, nn00008}@mix.wvu.edu, nasser.nasrabadi@mail.wvu.edu}
}

\maketitle
\thispagestyle{empty}
\pagestyle{empty}

\begin{abstract}
Deep convolutional neural networks have achieved remarkable success in face recognition (FR), partly due to the abundant data availability. However, the current training benchmarks exhibit an imbalanced quality distribution; most images are of high quality. This poses issues for generalization on hard samples since they are underrepresented during training. In this work, we employ the multi-model boosting technique to deal with this issue. Inspired by the well-known AdaBoost, we propose a sample-level weighting approach  to incorporate the importance of different samples into the FR loss. Individual models of the proposed framework are experts at distinct levels of sample hardness. Therefore, the combination of models leads to a robust feature extractor without losing the discriminability on the easy samples. Also, for incorporating the sample hardness into the training criterion, we analytically show the effect of sample mining on the important aspects of current angular margin loss functions, i.e., margin and scale. The proposed method shows superior performance in comparison with the state-of-the-art algorithms in extensive experiments on the CFP-FP, LFW, CPLFW, CALFW, AgeDB, TinyFace, IJB-B, and IJB-C evaluation datasets.

\end{abstract}

\section{Introduction}
The classical Face Recognition (FR) frameworks are based on extracting hand-crafted features \cite{ahonen2006face,facenet}. Nuisance factors such as head pose, resolution, blur, occlusion, and illumination variance in expression severely affect FR performance \cite{ahonen2006face,saadabadi2022information}. Since the advent of deep Convolutional Neural Networks (CNN) and the introduction of large-scale FR datasets, deep CNN-based FR have gained popularity \cite{schroff2015facenet,malakshan2023joint}.
The introduction of ResNet architecture and seminal works of \cite{VGGFace2,facenet}  have revolutionized FR into the challenge of finding robust and suitable loss functions \cite{he2016deep,deng2019arcface}. 
The general goal is to force the model to learn discriminative representations with a minimal intra-class distance and a maximal inter-class disparity \cite{liu2017sphereface}.
As a metric learning task, there are two main approaches to designing the loss function: 1) multi-class and 2) pair-wise supervision. 
\begin{figure}
\begin{center}
\includegraphics[width=0.8\linewidth]{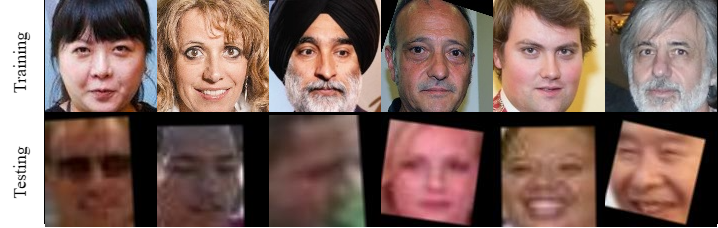}
\end{center}
   \caption{Illustrating the resolution and quality disparity between training (WebFace4M) and testing benchmarks (TinyFace).}
\label{tvst}
\vspace{-4mm}
\end{figure}

Due to the availability of large-scale labeled datasets, the regular choice of training objective is Softmax. However, as an open-set recognition task, the discriminability of feature representation matters \cite{wen2021sphereface2}. Despite being separable, the representation yielded through Softmax loss exhibits poor discriminability \cite{liu2017sphereface,wen2021sphereface2}. The pioneering works in [42, 54] improved discriminability by using deep metric learning loss functions. Most recently, using angular distance (instead of Euclidean) and angular penalty in the Softmax improved the discriminability power of feature representations \cite{saadabadi2023quality,liu2017sphereface,wang2018cosface,deng2019arcface}.
In the state-of-the-art (SOTA) deep FR framework, a model is trained using angular margin objective until convergence \cite{liu2017sphereface,wang2018cosface,deng2019arcface}. However, current SOTA performance demonstrates limited generalization on low-quality and Low-Resolution (LR) inputs, such as images captured by surveillance cameras or captured from long ranges \cite{zhang2021understanding}.
\begin{figure}
\begin{center}
\includegraphics[width=1.0\linewidth]{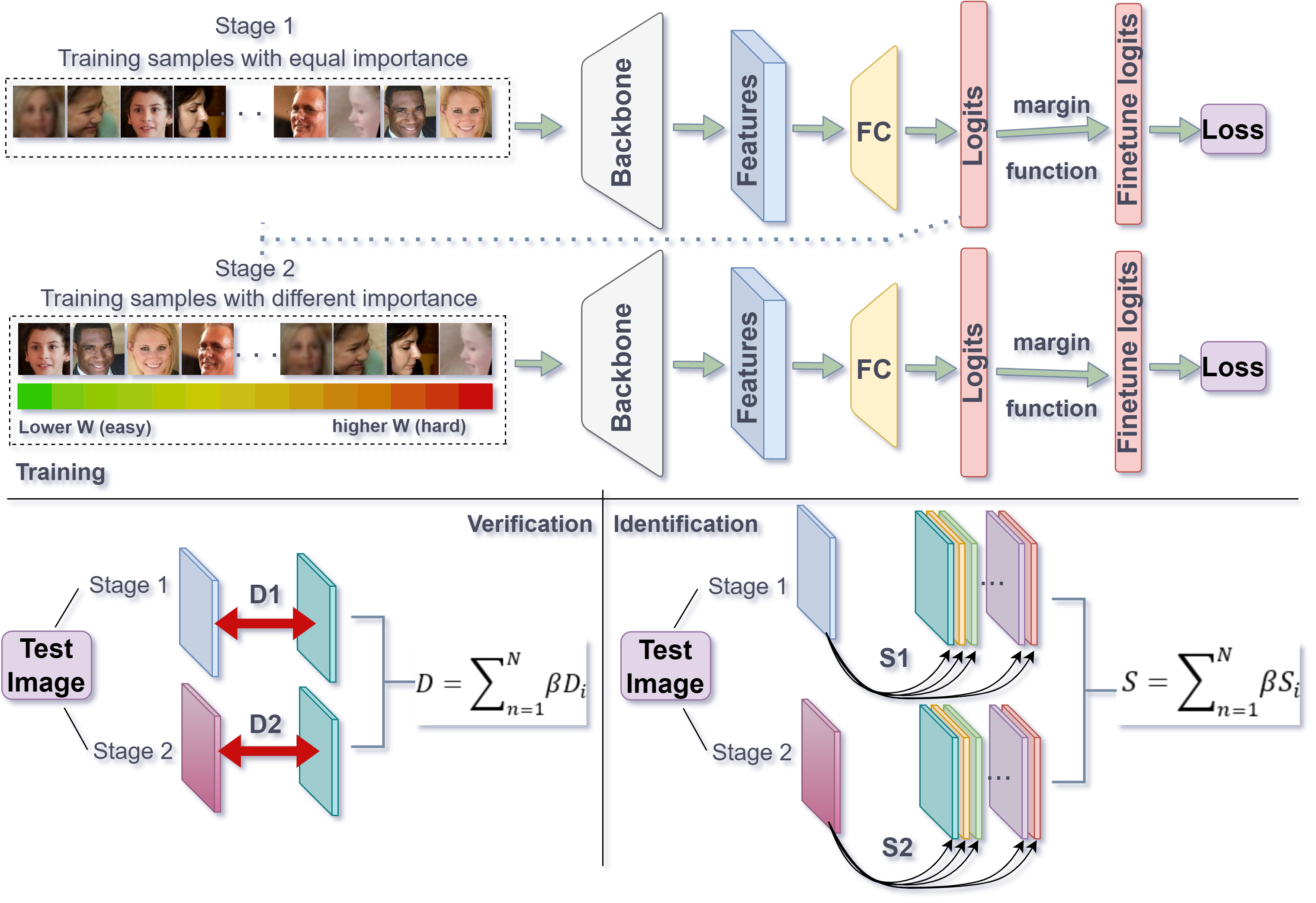}
\end{center}
   \caption{Schematic diagram of the proposed method, based on a CNN transfer learning for $K=1$ and $K=2$.}
\label{fig:overview}
\vspace{-4mm}
\end{figure}

Large-scale datasets such as WebFce260M \cite{zhu2021webface260m}, or MS-Celeb-1M \cite{guo2016ms}, mainly contain high-resolution (HR) instances that have significant statistical disparity from real-world surveillance images, as shown in Fig. \ref{tvst} \cite{hong2021stylemix}.
In other words, difficult samples, i.e., including LR instances, are under-represented. However, an implicit assumption in conventional angular margin methods is that all samples are equally important \cite{li2019airface,kim2022adaface}. Consequently, a manually selected margin squeezes all intra-class variations uniformly, which can be sub-optimal \cite{li2019airface,deng2019arcface,ren2018learning}. 
That being said, leveraging FR frameworks with sample hardness has attracted the researcher's attention to increase the model's discriminability power \cite{meng2021magface}.


\begin{figure*}
\begin{center}
\includegraphics[width=0.7\linewidth]{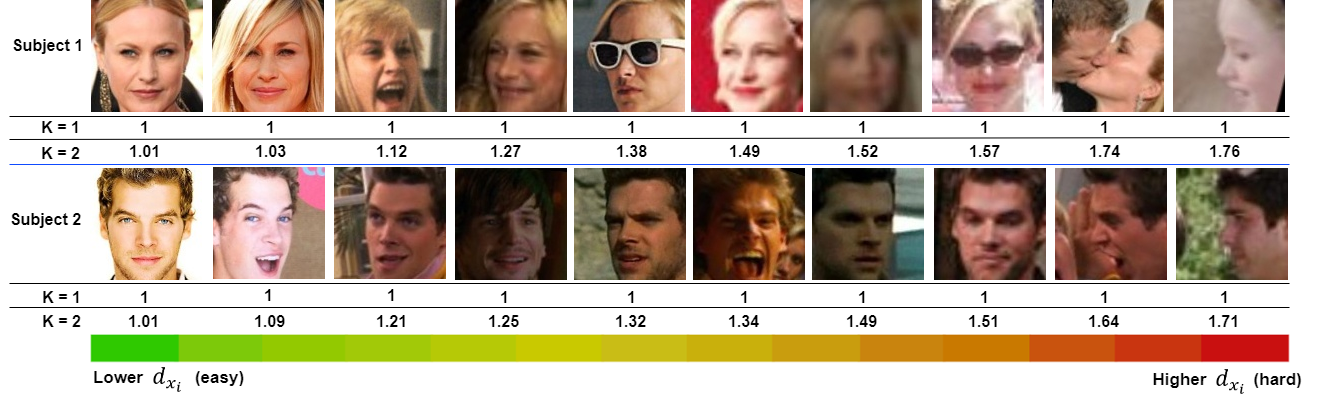}
\end{center}
   \caption{Various examples (including easy and hard samples) from two subjects and their assigned weights \(d_{x_{i}}\) by our method for ($K=1$) and ($K=2$).}

\label{fig:norm_analysis}
\vspace{-5mm}
\end{figure*}


Methods have been developed to de-emphasize or over-emphasize the instances based on selected characteristics as the proxy of sample hardness, such as feature norm or uncertainty \cite{wang2020mis,meng2021magface}. However, the performance improvement is inconsistent as they ignore hard samples \cite{li2019airface,kim2022adaface}. Also, convergence is not guaranteed due to the complex essence of the FR problem when solely trained on hard instances \cite{li2019airface}.
Data augmentation can enhance the frequency of difficult images and the diversity of training benchmarks \cite{shin2022teaching,kim2022adaface}. However, due to the tied angular margin in conventional FR methods, they suffer from convergence problems and cannot fit well with data augmentations \cite{zhang2023unifying}. 
Methods have been proposed to adaptively tune the margin based on the difficulty of the sample \cite{zhang2019adacos,liu2019adaptiveface,meng2021magface}. Although promising improvements have been gained via combining augmentations and adaptive margin, the performance still needs to be improved in testing benchmarks with a large distribution gap, implying that augmented samples cannot mimic the actual distribution of in-the-wild images.

An optimal FR model must generalize across different data distributions, such as off-angle, LR, and distorted images, to accommodate distribution shifts from training to real-world applications.
Ensemble learning is an effective method that trains a set of models on the whole or subsets of a dataset to mitigate the sensitivity to the distribution shift in the data \cite{opitz1999popular}.
Most recently, combining boosting strategies with CNNs for object classification and image denoising resulted in promising improvements \cite{han2016incremental,moghimi2016boosted,chen2018deep}. 
Boosting, as an ensemble learning technique,  is capable of generalizing over various distributions and good interpretability \cite{melville2005creating,jiang2018dimboost}. In contrast to methods that try to compensate for the scarcity of low-quality samples by introducing a sampling strategy \cite{ren2018learning,zhang2019adacos,liu2019adaptiveface,najafzadeh2023face,meng2021magface}, in boosting framework, distinct models are designed which are experts for samples with different hardness.
To achieve this, the optimization path changes in accordance with the sample's hardness, and as an ensemble of models, they can enhance the overall generalization power \cite{krogh1994neural}. 

We propose a method to take advantage of Adaptive Boosting (AdaBoost) to deal with the distribution shift on FR applications. To this aim, we hypothesize that in large-scale FR training benchmarks (i.e., LR, HR, long-range, and distorted images) are available (an imbalanced distribution concerning sample hardness). However, high-quality images dominate in these datasets (imbalance sample hardness).
We propose an ensemble learning framework such that each learner adjusts the weight of the training samples for the consecutive learners based on the samples' hardness. As a result, the subsequent learners will concentrate on distinct samples compared to their predecessors, which changes the optimization path and leads to a more diverse feature representation.
Consequently, we can: 1) explore currently available training data more effectively and 2) increase the generalizability of the resulting ensembled model on hard instances while maintaining the performance on easy facial images. 
Contributions of this work can be summarized as follows:

\begin{itemize}
  \item We move beyond class-level imbalance to propose a novel sample-level objective function inspired by AdaBoost that better compensates for data distribution imbalance and give more importance to the misclassified samples from the earlier trained model.

\item We empirically and theoretically show the relationship between sample mining and angular margin penalty.

\item We propose a method to relieve the convergence issue of the current FR training paradigm when there is more emphasis on hard samples.
\end{itemize}

\section{Related Work}

\subsection{Boosting}

Boosting has been used in ensemble models to enhance performance by cascading several sub-models \cite{han2016incremental,moghimi2016boosted,kashiani2022robust}. 
Out of the many methods used for boosting, AdaBoost and Gradient Boosting are two of the most commonly surfed techniques \cite{freund1996experiments,friedman2001greedy}. Boosting, also known as forward stagewise additive modeling, was originally proposed to improve the performance of classification trees. It has been recently incorporated into  deep learning models to improve their performance further. 
Schwenk and Bengio \cite{schwenk1997training} improved the ensemble accuracy of neural networks by using AdaBoost. Kawana et al. \cite{kawana2018ensemble} have proposed an ensemble of CNNs for human pose estimation, where each CNN in the ensemble model is optimized for a range of poses. They integrated the output of each individual CNN by feeding them as input to an integration module. Instead of averaging, the boosted CNN method in \cite{moghimi2016boosted} uses the least square objective function to incorporate the boosting weights into training. Since boosting increases the networks' complexity, dense connections were adopted in a deep boosting framework to tackle the problem of vanishing gradient \cite{chen2018deep}. Yang et al. \cite{yang2015convolutional} used CNN to generate high-level features, followed by a boosted Forest classifier. Boosting techniques have been studied in classical machine learning methods and some limited areas of deep learning; however, they have yet to be explored for deep FR.
\begin{table*}[]
\small
\begin{center}
\caption{Perfomance (\%) comparison of our method with other recent algorithms. 1:1 verification accuracy for LFW, CFP-FP, CPLFW, AgeDB, and closed-set rank retrieval for TinyFace are reported. The backbone used here is Resnet18.} \label{res18_highTiny}
\addtolength{\tabcolsep}{2pt}
\begin{tabular}{|l|llllll|lll|}
\hline
\multirow{2}{*}{Method} & \multicolumn{6}{c|}{High Quality}                                                                                                                        & \multicolumn{3}{c|}{Low Quality (TinyFace)}                                 \\ \cline{2-10} 
                        & \multicolumn{1}{l|}{LFW}    & \multicolumn{1}{l|}{CFP-FP} & \multicolumn{1}{l|}{CPLFW}  & \multicolumn{1}{l|}{CALFW}  & \multicolumn{1}{l|}{AgeDB}  & AVG             & \multicolumn{1}{l|}{Rank-1} & \multicolumn{1}{l|}{Rank-5} & Rank-20 \\ \hline
HM-Softmax    \cite{shrivastava2016training}          & \multicolumn{1}{l|}{97.77}       & \multicolumn{1}{l|}{90.11}       & \multicolumn{1}{l|}{83.25}       & \multicolumn{1}{l|}{89.55}       & \multicolumn{1}{l|}{90.23}       &    90.18             & \multicolumn{1}{l|}{32.21}       & \multicolumn{1}{l|}{37.55}       &    39.45     \\
MV-Softmax \cite{wang2020mis}             & \multicolumn{1}{l|}{98.25}       & \multicolumn{1}{l|}{91.36}       & \multicolumn{1}{l|}{84.47}       & \multicolumn{1}{l|}{91.88}       & \multicolumn{1}{l|}{91.15}       &       91.42          & \multicolumn{1}{l|}{36.19}       & \multicolumn{1}{l|}{47.14}       &   40.88      \\
CosFace      \cite{wang2018cosface}           & \multicolumn{1}{l|}{99.00}       & \multicolumn{1}{l|}{91.89}       & \multicolumn{1}{l|}{84.99}       & \multicolumn{1}{l|}{91.63}       & \multicolumn{1}{l|}{91.85}       &       91.87         & \multicolumn{1}{l|}{45.00}       & \multicolumn{1}{l|}{55.00}       &  58.00      \\
CurricularFace  \cite{huang2020curricularface}        & \multicolumn{1}{l|}{98.87}       & \multicolumn{1}{l|}{92.05}       & \multicolumn{1}{l|}{86.14}       & \multicolumn{1}{l|}{92.46}       & \multicolumn{1}{l|}{92.24}       &    92.35             & \multicolumn{1}{l|}{45.15}       & \multicolumn{1}{l|}{48.14}       &    54.24     \\
ArcFace     \cite{deng2019arcface}            & \multicolumn{1}{l|}{99.01}       & \multicolumn{1}{l|}{92.76}       & \multicolumn{1}{l|}{86.16}       & \multicolumn{1}{l|}{92.65}       & \multicolumn{1}{l|}{92.70}       &    92.65             & \multicolumn{1}{l|}{52.47}       & \multicolumn{1}{l|}{58.63}       &      62.23   \\
AdaFace   \cite{kim2022adaface}              & \multicolumn{1}{l|}{99.13} & \multicolumn{1}{l|}{92.82} & \multicolumn{1}{l|}{87.00}  & \multicolumn{1}{l|}{92.65}  & \multicolumn{1}{l|}{92.717} & 92.86          & \multicolumn{1}{l|}{56.06} & \multicolumn{1}{l|}{61.45} & 65.21  \\
\hline
Ours (K=1 \& 2 )         & \multicolumn{1}{l|}{ \textbf{99.23} }& \multicolumn{1}{l|}{ \textbf{92.96} }& \multicolumn{1}{l|}{ \textbf{87.07} }& \multicolumn{1}{l|}{ \textbf{92.93}} & \multicolumn{1}{l|}{ \textbf{93.00}}  &  \textbf{93.04        } & \multicolumn{1}{l|}{ \textbf{57.83}} & \multicolumn{1}{l|}{ \textbf{63.31}} &  \textbf{67.17} \\

\hline
\end{tabular}
\label{table:high&tinyRes18}
\vspace{-6mm}
\end{center}

\end{table*}


 \subsection{Margin-based Face Recognition}
In recent years, most of the studies in the area of deep FR have been dedicated to enhancing performance by devising novel training criteria.
The standard Softmax loss does not provide sufficient discriminability in embedding, i.e., intra-class compactness and inter-class separability \cite{he2020softmax}. The pioneer works of FaceNet \cite{facenet} introduced a novel loss function that simultaneously uses positive (same identities) and negative (different identities) samples to improve the deep representations' discriminative power. This is achieved by bringing the anchor and positive sample closer together in the embedding space and pushing the anchor away from the negative sample \cite{wen2016discriminative,khosravi2022improved,sun2014deep}.
Several studies on the characteristics of the Softmax embeddings found an angular distribution in the representations. Therefore, the most recent methods proposed to increase the discriminability power of feature representation by mapping the Euclidian similarity of standard Softmax to angular space. As a result, by incorporating the angular penalty into the angular Softmax function, SOTA results have been achieved in various studies, including CosFace \cite{wang2018cosface}, ArcFace \cite{deng2019arcface}, AdaFace \cite{kim2022adaface} and QAFace \cite{saadabadi2023quality} .

\subsection{Sample Mining in Face Recognition}

Improving generalization ability is essential, and one way to achieve this is through hard sample mining \cite{liu2022learning}. Studies in this area have focused on two aspects: 1) finding a measure of sample hardness and 2) incorporating the sample hardness into the training paradigm \cite{wang2020mis}.
Shrivastava et al., in \cite{shrivastava2016training}, find easy and hard samples based on the loss value. They emphasize hard samples and discard easy samples (HM-SoftMax) to improve the generalization. Lin et al. \cite{lin2017focal} re-weights all the samples by introducing a soft mining strategy and training the network on a sparse set of hard instances \cite{lin2017focal}. Recently, MV-Softmax \cite{wang2020mis} has emerged as a framework that integrates margin and mining techniques. This approach defines hard samples as misclassified and emphasizes them by applying a predetermined constant on their negative cosine similarities.
CurricularFace \cite{huang2020curricularface} employs the Curriculum Learning strategy to focus on easy samples at the beginning of training and then shifts the emphasis toward hard instances. Furthermore, Liu et al. \cite{liu2022learning} showed that samples within the same class have varying levels of importance and employed meta-learning to assign weights to each sample based on multiple variation factors. They trained a model with four learnable margins corresponding to ethnicity, pose, blur, and occlusion to achieve this. One major shortcoming of these works is the inconsistency of the improvements, i.e., increasing the performance on the harder benchmarks is gained by sacrificing performance on the easy benchmarks \cite{saadabadi2023quality}. We seek to train multiple agents, e.g., deep CNNs, and combine their output to obtain consistency across different benchmarks. Also, we strive to mitigate the challenge of finding the hardness measure by directly employing the model score to indicate hardness.


\section{Proposed Work}

\subsection{Overview of Angular FR Objective }

The standard deep recognition training framework consists of a stack of non-linear feature extractor layers (backbone) followed by a classifier \cite{boutros2021mfr,abbasian2023controlling}. The whole architecture is trained using gradient descent with angular penalty criterion:

\begin{equation}\label{angularmarginloss}
 \small
 \begin{aligned}
		L= -\frac{1}{N}\sum_{i=1}^{N}d_{x_i}{\log{\frac{e^{f(W_{y_i},x_i,M)}}{{e^{f(W_{y_i},x_i,M)}}+{\sum_{\substack{j=1 \\ j \neq y_i}}^{C}{e^{f(W_{j},x_i,M)}}}}}},
\end{aligned}
\end{equation}
where $W_j \in \mathbb{R}^{dim}$ is $j$-th classifier (center), and $dim$ is the feature dimension, $x_i$ is the learned feature of $i$-th sample, and $y_i$ is its corresponding ground truth. $N$ and $C$ represent the mini-batch size and the total number of classes, respectively. $M=(m_s,m_c,m_a)$ is the margin hyperparameter, $f$ is a function of $W_j$, $x_i$, and margin. $d_{x_i}$ is the indicator function which in angular margin losses is chosen to be one, i.e., the equal importance of samples. Usually, when the feature vectors and class centers are projected to the unit-hypersphere, then $f$ is written as a function of the angle between the feature vector and the $j$-th center of the classifier, $\small{f(W_{j},x_i,M)=f(\theta_{j,i},M)}$:
\begin{equation}\label{margin_function}
 \small
 \begin{aligned}
 & \\
 f(\theta_{j,i},M)=&\left\{ 
  \begin{array}{ l l }
    s\:cos(m_s\theta_{j,i}+m_a)-m_c;& j=y_i,  \\
    s\:cos(\theta_{j,i});&j\neq y_i,
  \end{array}
 \right.
\end{aligned}
\end{equation}
where the $\theta_{j,i}$ represents the angle between $j$-th center and $i$-th sample.

\subsection{Integrating Ensemble Boosting to FR Training}

\begin{figure}
\begin{center}
\includegraphics[width=0.7\linewidth]
{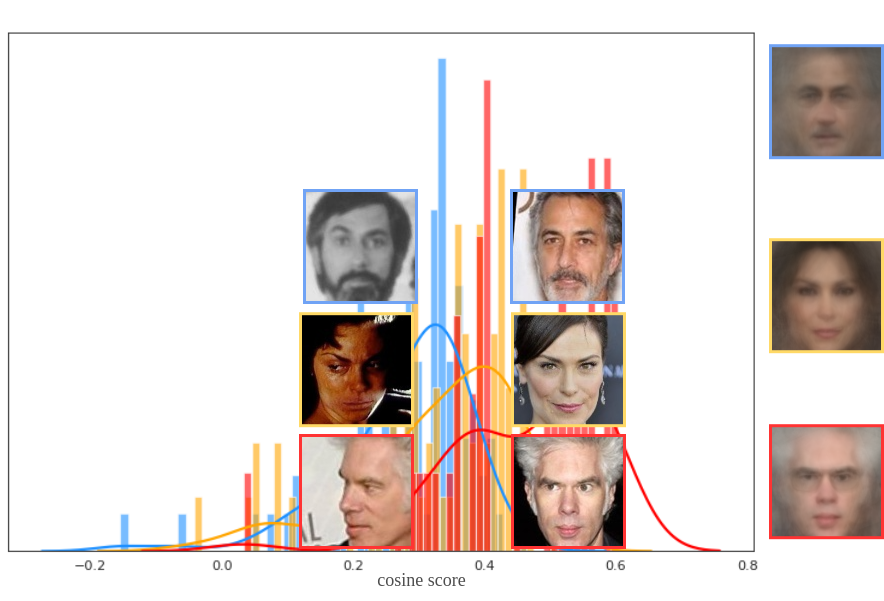}
\vspace{-6mm}
\end{center}
   \caption{Similarity scores between samples in each class and its corresponding class center (average of samples), higher scores represent the high-quality and most frequent samples and lower scores represent the low-quality and hard samples.}
\label{simwithmean}
\vspace{-4mm}
\end{figure}

AdaBoost was initially designed for a binary classification task in combination with a decision-tree algorithm \cite{freund1995desicion}. Here, we utilize its original idea to fit it in a multi-class paradigm of FR training \cite{hastie2009multi}.
The goal is to enhance the representation power on the hard instances while maintaining the performance on the easy images. To do so, we consider K models (K=2 in our experiments) to form our ensemble model. The first model should be trained using the standard FR framework, $d_{x_i} =1$ in Eq.~\ref{angularmarginloss}. 
The classifier's centers can be considered the average of the samples in each class \cite{saadabadi2023quality}. Therefore, Eq.~\ref{margin_function} reflects the similarity of each sample with the average of samples in the specified class. Because the FR training benchmarks are imbalanced concerning the hardness of the samples \cite{kim2022adaface}, class centers tend to drift toward more frequent samples, which results in reducing the loss by increasing the similarity between the centers and over-represented instances, see Fig.~\ref{simwithmean}. Consequently, the model is inclined toward forgetting the hard instances during the training, which can result in a lack of generalization over the hard samples.

 \begin{table*}[]
\small
\begin{center}
\caption{Perfomance (\%) comparison of our method with other recent algorithms. True Acceptance Rate (TAR) at a different level of False Acceptance Rate (FAR) are reported for IJB-B and IJB-C. The backbone used here is Resnet18.} \label{res18IJB}
\addtolength{\tabcolsep}{0.75pt}
\begin{tabular}{|l|cccccc|cccccc|}
\hline
\multirow{2}{*}{Method} & \multicolumn{6}{c|}{Mix Quality (IJB-B)}                                                                                                                             & \multicolumn{6}{c|}{Mix Quality (IJB-C)}                                                                                                                          \\ \cline{2-13} 
                        & \multicolumn{1}{c|}{1e-06} & \multicolumn{1}{c|}{1e-05} & \multicolumn{1}{c|}{1e-04} & \multicolumn{1}{c|}{0.001} & \multicolumn{1}{c|}{0.01}  & 0.1   & \multicolumn{1}{c|}{1e-06} & \multicolumn{1}{c|}{1e-05} & \multicolumn{1}{c|}{1e-04} & \multicolumn{1}{c|}{0.001} & \multicolumn{1}{c|}{0.01} & 0.1 \\ \hline
HM-Softmax    \cite{shrivastava2016training}          & \multicolumn{1}{c|}{0.00}      & \multicolumn{1}{c|}{0.00}      & \multicolumn{1}{c|}{8.06}      & \multicolumn{1}{c|}{68.80}      & \multicolumn{1}{c|}{87.07}      &    93.12   & \multicolumn{1}{c|}{0.00}      & \multicolumn{1}{c|}{0.10}      & \multicolumn{1}{c|}{8.76}      & \multicolumn{1}{c|}{64.37}      & \multicolumn{1}{c|}{88.68}     & 90.76     \\
MV-Softmax  \cite{wang2020mis}            & \multicolumn{1}{c|}{0.00}      & \multicolumn{1}{c|}{0.00}      & \multicolumn{1}{c|}{0.00}      & \multicolumn{1}{c|}{8.90}      & \multicolumn{1}{c|}{68.97}      &    94.11   & \multicolumn{1}{c|}{0.00}      & \multicolumn{1}{c|}{0.15}      & \multicolumn{1}{c|}{10.54}      & \multicolumn{1}{c|}{68.15}      & \multicolumn{1}{c|}{90.05}     &   94.60  \\
CosFace   \cite{wang2018cosface}              & \multicolumn{1}{c|}{0.00}      & \multicolumn{1}{c|}{0.11}      & \multicolumn{1}{c|}{10.01}      & \multicolumn{1}{c|}{70.16}      & \multicolumn{1}{c|}{89.69}      &    95.81   & \multicolumn{1}{c|}{0.00}      & \multicolumn{1}{c|}{0.80}      & \multicolumn{1}{c|}{14.75}      & \multicolumn{1}{c|}{69.05}      & \multicolumn{1}{c|}{91.58}     &  96.02   \\
CurricularFace     \cite{huang2020curricularface}     & \multicolumn{1}{c|}{0.00}      & \multicolumn{1}{c|}{0.15}      & \multicolumn{1}{c|}{10.14}      & \multicolumn{1}{c|}{70.95}      & \multicolumn{1}{c|}{90.01}      &    95.86   & \multicolumn{1}{c|}{0.01}      & \multicolumn{1}{c|}{1.01}      & \multicolumn{1}{c|}{15.25}      & \multicolumn{1}{c|}{69.18}      & \multicolumn{1}{c|}{91.89}     & 96.42    \\
ArcFace       \cite{deng2019arcface}          & \multicolumn{1}{c|}{0.01}      & \multicolumn{1}{c|}{1.02}      & \multicolumn{1}{c|}{12.27}      & \multicolumn{1}{c|}{71.89}      & \multicolumn{1}{c|}{91.98}      &   97.11    & \multicolumn{1}{c|}{0.13}      & \multicolumn{1}{c|}{1.12}      & \multicolumn{1}{c|}{15.98}      & \multicolumn{1}{c|}{70.19}      & \multicolumn{1}{c|}{93.41}     & 97.83     \\
AdaFace      \cite{kim2022adaface}           & \multicolumn{1}{c|}{0.11}  & \multicolumn{1}{c|}{1.26}  & \multicolumn{1}{c|}{13.28} & \multicolumn{1}{c|}{72.81} & \multicolumn{1}{c|}{92.53} & 97.40 & \multicolumn{1}{c|}{0.13}      & \multicolumn{1}{c|}{1.26}      & \multicolumn{1}{c|}{17.80}      & \multicolumn{1}{c|}{71.23}      & \multicolumn{1}{c|}{93.44}     &   97.84  \\ \hline
Ours (K=1 \& 2 )         & \multicolumn{1}{c|}{ \textbf{1.70}}  & \multicolumn{1}{c|}{ \textbf{7.20} } & \multicolumn{1}{c|}{ \textbf{40.63}} & \multicolumn{1}{c|}{ \textbf{82.55}} & \multicolumn{1}{c|}{ \textbf{93.36}} &  \textbf{97.88} & \multicolumn{1}{c|}{ \textbf{1.14}}      & \multicolumn{1}{c|}{ \textbf{6.12}}      & \multicolumn{1}{c|}{\textbf{41.42}}      & \multicolumn{1}{c|}{ \textbf{83.37}}      & \multicolumn{1}{c|}{ \textbf{94.68}}     &     \textbf{98.25}
\\ \hline
\end{tabular}
\label{table:IJB}
\vspace{-5mm}
\end{center}
\end{table*}

For a given sample $i$, Eq.~\ref{margin_function} reflects the similarity of the sample with its class center, which can be interpreted as the sample's hardness \cite{meng2021magface}. To obtain the samples' hardness, between the range 0 and 1, we explicitly utilize the Softmax output score instead of the Cosine score:
\begin{equation}\label{prob}
 \small
 \begin{aligned}
		p_i= \frac{e^{f(W_{y_i},x_i,M)}}{{e^{f(W_{y_i},x_i,M)}}+{\sum_{\substack{j=1 \\ j \neq y_i}}^{C}{e^{f(W_{j},x_i,M)}}}}.
\end{aligned}
\end{equation}
\noindent In the standard AdaBoost framework, the constraint is that each binary classifier's accuracy is better than random guessing rather than 1/2. However,  we want to increase the discrimination power among the hard samples in the FR training paradigm. It is important to mention that solely training on the hard instances can lead to suboptimal solutions and overfitting \cite{opitz2017bier}. Our approach obtains sample weights independent from the visual quality, such as feature norm, and the hardness has directly resulted from the optimization path. To this aim, we chose a simple yet effective weighting scheme that is easy to implement and comprehend, as given by Eq. \ref{weight}: 
\begin{equation}\label{weight}
 \small
 \begin{aligned}
{d_{x_i}}^{k+1}={d_{x_i}}^k	p_i^{-\alpha}.
\end{aligned}
\end{equation}

\begin{figure}
\begin{center}
\includegraphics[width=0.7\linewidth]{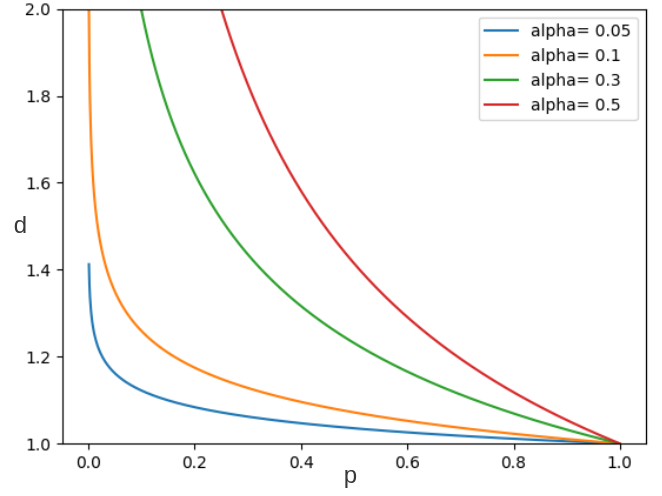}
\vspace{-4mm}
\end{center}
   \caption{Illustrating the proposed sample hardness, $d$, against the output probability of the classifier. It is important to consider the impact of the parameter $\alpha$ in Eq. \ref{weight}. When $\alpha$ has a large value, it can result in the weights of certain samples being overemphasized and easier samples being disregarded; more information is in Table \ref{ablation_alpha}.}
\label{fig:dvsp}
\vspace{-4mm}
\end{figure}

\noindent Eq.~\ref{weight} puts more emphasis on the hard instances in a way that more challenging samples in each class will receive more weights during training, as shown in Fig.~\ref{fig:dvsp}. Also, since the easy samples are not completely ignored, they relieve the feature representations from collapsing \cite{robbins2022effect}.
In evaluation, the final matching score is the weighted sum over the match score of K backbones:

\begin{table*}[]
\small
\begin{center}
\caption{Perfomance (\%) comparison of our method with other recent algorithms. 1:1 verification accuracy for LFW, CFP-FP, CPLFW, AgeDB, closed-set rank retrieval for TinyFace and TAR@FAR=0.01\% for IJB-B and IJC-B are reported. The backbone used here is Resnet50.}\label{high&tinyRes50}
\addtolength{\tabcolsep}{-0.5pt}
\begin{tabular}{|l|llllll|lll|ll|}

\hline
\multirow{2}{*}{Method} & \multicolumn{6}{c|}{High Quality}                                                                                                                        & \multicolumn{3}{c|}{Low Quality (TinyFace)} & \multicolumn{2}{c|}{Mix Quality}                                \\ \cline{2-12} 
                        & \multicolumn{1}{l|}{LFW}    & \multicolumn{1}{l|}{CFP-FP} & \multicolumn{1}{l|}{CPLFW}  & \multicolumn{1}{l|}{CALFW}  & \multicolumn{1}{l|}{AgeDB}  & AVG             & \multicolumn{1}{l|}{Rank-1} & \multicolumn{1}{l|}{Rank-5} & Rank-20& \multicolumn{1}{l|}{IJB-B} & IJB-C \\ \hline
HM-Softmax      \cite{shrivastava2016training}        & \multicolumn{1}{l|}{97.85}       & \multicolumn{1}{l|}{92.85}       & \multicolumn{1}{l|}{90.14}       & \multicolumn{1}{l|}{91.75}       & \multicolumn{1}{l|}{92.33}       &         92.98        & \multicolumn{1}{l|}{46.71}       & \multicolumn{1}{l|}{48.21}       &    50.47 & \multicolumn{1}{l|}{89.10}       &    62.96     \\
MV-Softmax        \cite{wang2020mis}      & \multicolumn{1}{l|}{99.08}       & \multicolumn{1}{l|}{94.39}       & \multicolumn{1}{l|}{93.10}       & \multicolumn{1}{l|}{94.01}       & \multicolumn{1}{l|}{92.33}       &        94.58         & \multicolumn{1}{l|}{52.36}       & \multicolumn{1}{l|}{55.74}       &     58.89 & \multicolumn{1}{l|}{91.39}       &     64.14   \\
CosFace    \cite{wang2018cosface}             & \multicolumn{1}{l|}{99.51}       & \multicolumn{1}{l|}{95.44}       & \multicolumn{1}{l|}{93.90}       & \multicolumn{1}{l|}{94.70}       & \multicolumn{1}{l|}{94.56}       &         95.62        & \multicolumn{1}{l|}{60.14}       & \multicolumn{1}{l|}{63.77}       &     65.77& \multicolumn{1}{l|}{92.52}       &     65.42    \\
CurricularFace   \cite{huang2020curricularface}       & \multicolumn{1}{l|}{99.42}       & \multicolumn{1}{l|}{96.32}       & \multicolumn{1}{l|}{93.85}       & \multicolumn{1}{l|}{94.78}       & \multicolumn{1}{l|}{94.81}       &        95.84         & \multicolumn{1}{l|}{61.89}       & \multicolumn{1}{l|}{65.51}       & 67.86& \multicolumn{1}{l|}{92.51}       & 65.26        \\
ArcFace       \cite{deng2019arcface}          & \multicolumn{1}{l|}{99.70}       & \multicolumn{1}{l|}{97.14}       & \multicolumn{1}{l|}{94.05}       & \multicolumn{1}{l|}{95.14}       & \multicolumn{1}{l|}{95.15}       &         96.24        & \multicolumn{1}{l|}{68.99}       & \multicolumn{1}{l|}{73.89}       &    76.04 &\multicolumn{1}{l|}{94.39}       &    96.19     \\
AdaFace      \cite{kim2022adaface}           & \multicolumn{1}{l|}{ \textbf{99.78}} & \multicolumn{1}{l|}{97.14} & \multicolumn{1}{l|}{ \textbf{94.16}}  & \multicolumn{1}{l|}{95.98}  & \multicolumn{1}{l|}{97.78} & 96.97          & \multicolumn{1}{l|}{70.25} & \multicolumn{1}{l|}{74.034} & 76.31& \multicolumn{1}{l|}{95.44} & 96.98  \\
\hline
Ours (K=1 \& 2)         & \multicolumn{1}{l|}{99.75}  & \multicolumn{1}{l|}{ \textbf{97.24}} & \multicolumn{1}{l|}{94.13} & \multicolumn{1}{l|}{ \textbf{96.05}} & \multicolumn{1}{l|}{ \textbf{97.85}}  &  \textbf{97.01} & \multicolumn{1}{l|}{ \textbf{71.01}} & \multicolumn{1}{l|}{ \textbf{74.54}} &  \textbf{76.80} & \multicolumn{1}{l|}{ \textbf{95.47}} &   \textbf{97.00} \\

\hline
\end{tabular}
\vspace{-4mm}
\end{center}

\end{table*}


\begin{equation}\label{test}
 \small
 \begin{aligned}
H_{final}(x_i)=\sum_{k=1}^{K} \beta_k H_k(x_i),
\end{aligned}
\end{equation}
where $\beta_{k}$ hyperparameters are the weights associated to each trained model.

\subsection{Sample Hardness as Angular Margin}

There are two categories of sample mining methods: over-sampling and weighting schemes \cite{wang2020mis}. In the context of FR, over-sampling can lead to poor performance by reducing the diversity of samples. To alleviate this issue, one may resort to weighting methods \cite{zhang2019adacos,wang2020mis}.
Here, we propose that naively applying samples' weights to the angular framework can be suboptimal. Although introducing sampling importance aims to compensate for misclassified data, sample weights can have hidden affect in the angular space. Applying sample mining, the term $d_{x_i}$ in Eq. \ref{angularmarginloss} is no longer uniform for all the samples. 

\begin{figure}
\begin{center}
\includegraphics[width=0.7\linewidth]{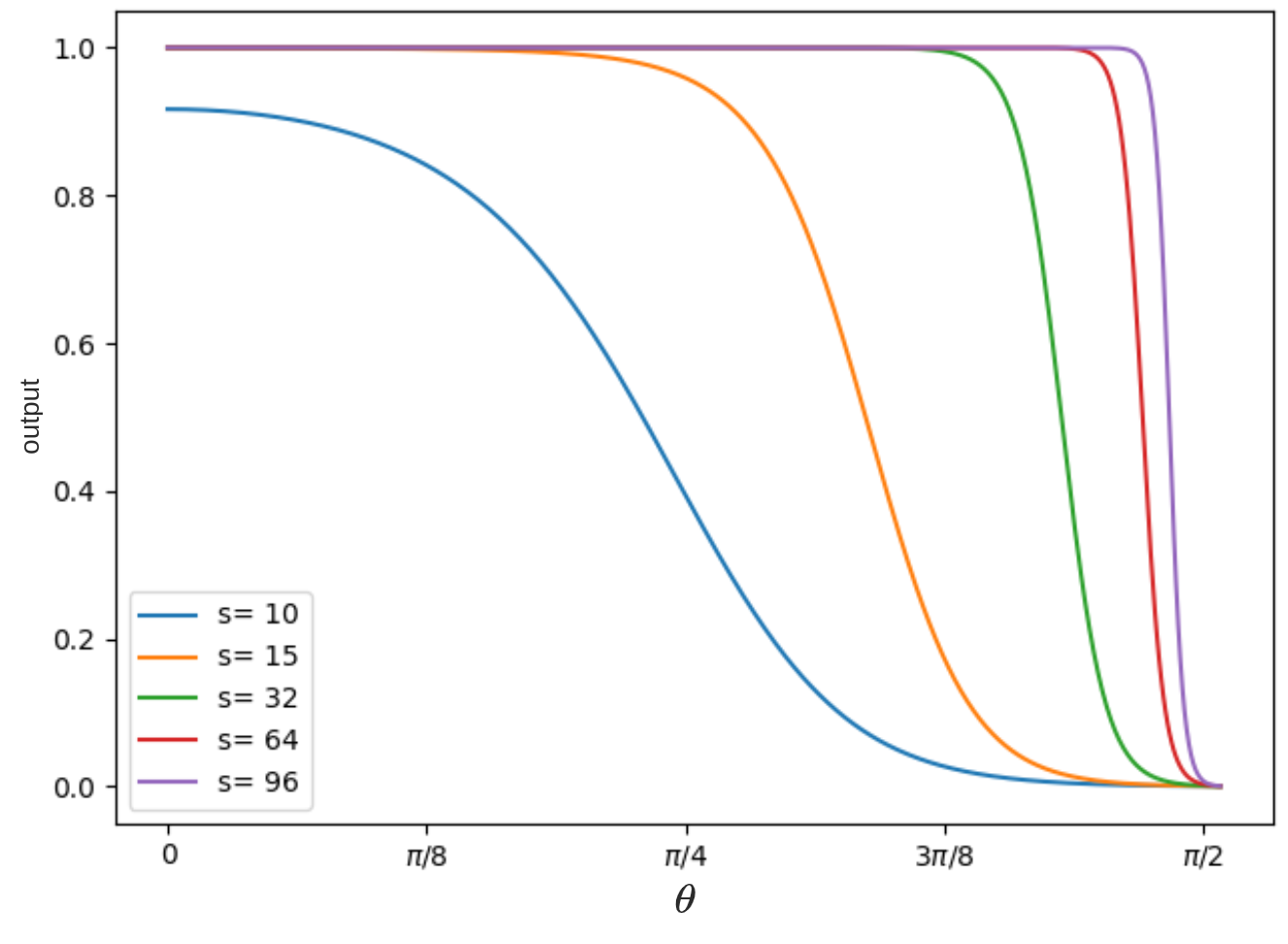}
\vspace{-4mm}
\end{center}
   \caption{Curves of Softmax output w.r.t. $\theta_i$ by choosing different scale values.}\label{seffect}
\vspace{-4mm}
\end{figure}

In Eq. \ref{angularmarginloss}, as the training converges, the denominator can be estimated by a constant value \cite{zhang2019adacos}. In convergence, cosine similarity between samples and negative classes is close to zero, i.e., $f(W_{j},x_i)= f(\theta_{j,i})= 0$. Therefore, the denominator can be approximated as: 
\begin{equation}\label{denomapp}
 \small
 \begin{aligned}
		e^{f(W_{y_i},x_i,M)}+{\sum_{\substack{j=1 \\ j \neq y_i}}^{C}{e^{f(W_{j},x_i,M)}}}  \approx e^{f(W_{y_i},x_i,M)} + C-1,
\end{aligned}
\end{equation}
replacing the denominator in Eq.~\ref{angularmarginloss} with Eq.~\ref{denomapp}, then we can rearrange the Eq.~\ref{angularmarginloss} as: 

\begin{equation}\label{angularmarginlossv2}
 \small
 \begin{aligned}
		L= -\frac{1}{N}\sum_{i=1}^{N}{\log{\left(\frac{e^{f(W_{y_i},x_i,M)}}{{e^{f(W_{y_i},x_i,M)}}+C-1}\right)^{d_{x_i}}}},
\end{aligned}
\end{equation}
consequently, the essential and differentiable component is the $f(W_{y_i},x_i,M)^{d_{x_i}}$ which can be rewriten as the following: 
\begin{equation}\label{d2margin}
 \small
 \begin{aligned}
\left(e^{f(W_{y_i},x_i,M)}\right)^{d_{x_i}} & = \left(e^{f(\theta_j,M)}\right)^{d_{x_i}} \\ &
= \left(e^{{d_{x_i}}\:f(\theta_j,M)}\right) \\ &
= \left(e^{s\: \cdot {d_{x_i}}\:cos(m_s\theta_j+m_a)-{d_{x_i}}\: m_c}\right) .
\end{aligned}
\end{equation}
In other words, the scaling factor and the cosine margin are adaptively tuned with respect to the samples' hardness. 
To further study the effect of the scale hyperparameter, $s$, we plot the Softmax output score versus $\theta_{i,j}$ for different $s$ as shown in Fig. \ref{seffect}. When the value of $s$ is too small, such as $s = 10$, it is apparent that the maximum output score cannot reach one. This outcome is not desirable because even if the network is highly confident in the corresponding prediction, the loss function will still penalize the network, leading to poor performance for easy samoples. Conversely, when $s$ is excessively large, the output curve is problematic as it produces a very high probability even when the angle is close to $\frac{\pi}{2}$. Consequently, the loss function with large $s$ may not penalize misclassified samples, resulting in poor performance on hard samples.
To alleviate this issue, we propose to tune the scale value with the normalized hardness score: 
\begin{equation}\label{dnormalized}
 \small
 \begin{aligned}
 \widehat{d}= \frac{d-\overline{d}}{std(d)} \: \lambda,
\end{aligned}
\end{equation}
where $\overline{d}$ is the moving average over the seen samples' weights, and $std$ represents the standard deviation. $\small{\frac{d-\overline{d}}{std(d)}}$ which makes the batch distribution of $d$ close to the normal Gaussian with zero mean and unit standard deviation. To further increase the concentration around the zero, $\lambda$ is used as a hyperparameter. Then, we use the $\widehat{d}$ to fine-Ftune the scale, $s$:
\begin{equation}\label{scaletune}
 \small
 \begin{aligned}
 s'=s-\lfloor \widehat{d} \rceil_{-0.33}^{0.33}\: s ,
\end{aligned}
\end{equation}
we clip the $\widehat{d}$ to be within $(-0.33,0.33)$ so the noisy samples do not distract the training \cite{kim2022adaface}. 
By this adaptivity, we can emphasize the hard samples while maintaining the discrimination on the easy instances \cite{liu2022learning}. A low value of $d$ results in negative $\widehat{d}$ which increases the value of $s$ (higher output score for easy samples) and vice versa. 


\section{Experimental Results}

\subsection{Datasets}
We employ publicly available WebFace4M \cite{zhu2021webface260m} as our training datasets which is a subset of the recently released FR dataset called WebFace260M. WebFace4M contains around 4M samples from 200k identities.
Following the protocol of \cite{shi2020towards}, we evaluate our models on five widely applied benchmarks in good quality, including LFW \cite{huang2008labeled}, CFP-FP \cite{sengupta2016frontal}, CPLFW \cite{zheng2018cross}
AgeDB \cite{moschoglou2017agedb} and CALFW \cite{zheng2017cross}. Also, two mixed-quality datasets from the Janus program, including: 
the IARPA Janus Benchmark-B (IJB-B) \cite{whitelam2017iarpa} and Benchmark-C (IJB-C) \cite{maze2018iarpa} were used in our evaluations. Additionally, we use TinyFace as a challenging low-quality evaluation benchmark \cite{cheng2019low}. 

\noindent{\textbf{IJB-B and IJB-C}}: The IJB-B \cite{whitelam2017iarpa} dataset is a collection of face images and videos that is used to benchmark FR systems. The experimental protocols for IJB-B follow the standard 1:1 verification protocol. A template-based matching process is used, where the global feature vector for each template is obtained by averaging over the instances in the template. IJB-C is an extension of the IJB-B dataset. The testing protocol for IJB-C is similar to the protocol for IJB-B.

\begin{figure}
\begin{center}
\includegraphics[width=0.6\linewidth]{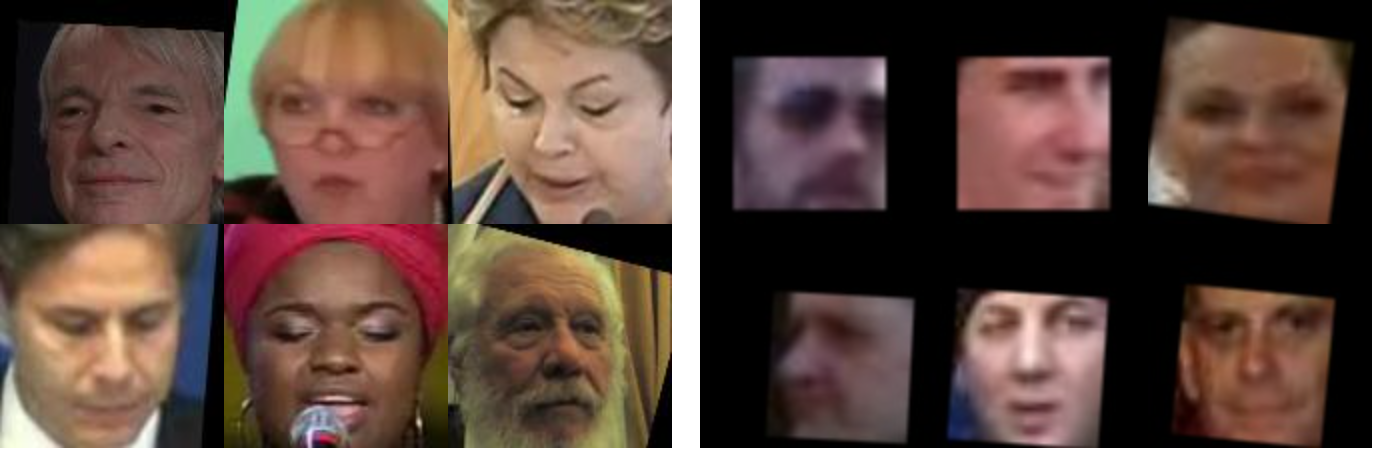}
\vspace{-4mm}
\end{center}
   \caption{Left: samples from the IJB-B dataset. Right: sample from TinyFace. IJB-B consists of high-quality and low-quality images, while TinyFace mainly contains low-quality samples. }
\label{tinyijb}
\vspace{-6mm}
\end{figure}

\noindent{\textbf{TinyFace}}: The TinyFace is a low-quality FR evaluation dataset. The images are designed for 1:$N$ recognition tests and have an average size of 20$\times$16 pixels. They were collected from public web data and face were captured under various uncontrolled conditions, including different poses, illumination, occlusion, and backgrounds. Fig. \ref{tinyijb} illustrates the quality of some samples from TinyFace and IJB-B datasets.


\subsection{Metrics}
There are two main ways to evaluate the performance of a face recognition paradigm: recognition and verification. Recognition is a 1:N task where the network calculates the similarity score of a given probe image against all the samples in a gallery and identifies the probe image. Verification is a 1:1 task in which the network determines whether a given pair of images represents the same identity. We report the verification results on the LFW, CFP-FP, AgeDB, CPLFW, IJB-B, IJB-C, and CALFW datasets. The identification results are reported on the TinyFace dataset.

\subsection{Implementation}
 
We followed the ArcFace setup for preprocessing \cite{deng2019arcface}. All the images are resized to 112\(\times\)112, aligned canonical view, and pixel values are normalized to \([-1,1]\). The experiments are conducted with ResNet18 and Resnet50 as the backbone, and the models, are trained for 24 epochs with AdaFace loss \cite{kim2022adaface}. The optimizer is SGD, with the learning rate starting from 0.1, which is decreased by a factor of 10 at epochs \{10, 16, 22\}. The optimizer weight-decay is set to 0.0001, the mini-batch size on each GPU is 512, and the model is trained using two Quadro RTX 8000.
Fig. \ref{fig:overview} shows the architecture and our proposed method for $K=1$ and $K=2$ in both training and evaluation settings. We investigate the impact of varying values of $\alpha$ in Eq. \ref{weight} on the performance of our method across different evaluation datasets. Our empirical analysis in Table \ref{ablation_alpha} showed that an alpha value of $0.1$ produced the best results. Also, $\beta_1=1$ and $\beta_2=0.1$ are obtained empirically. 

\begin{table*}[]
\begin{center}
\caption {Ablation study on model's performance for $K=1$, $K=2$, and the combination of them in different settings. OursV1 (best): using all the training samples with assigned weights for training the second model ($K=2$); OursV2: using hard samples for training the second model and OursV3: training the second model ($K=2$) from the scratch.}
\label{compare_ablation}
\addtolength{\tabcolsep}{-0.5pt}

\begin{tabular}{|l|llllll|ll|ll|}
\hline
\multicolumn{1}{|c|}{\multirow{2}{*}{Method}} & \multicolumn{6}{c|}{High Quality}                                                                                                                            & \multicolumn{2}{c|}{TinyFace} & \multicolumn{2}{l|}{Mixed Quality} \\ \cline{2-11} 
\multicolumn{1}{|c|}{}                        &  \multicolumn{1}{l|}{LFW}    & \multicolumn{1}{l|}{CFP-FP} & \multicolumn{1}{l|}{CPLFW}  & \multicolumn{1}{l|}{CALFW}  & \multicolumn{1}{l|}{AgeDB}  & AVG    & \multicolumn{1}{l|}{Rank-1}    & Rank-5     & \multicolumn{1}{l|}{IJB-B} & IJB-C \\ \hline
AdaFace (K=1)                                       & \multicolumn{1}{l|}{99.13} & \multicolumn{1}{l|}{92.83} & \multicolumn{1}{l|}{87.00}  & \multicolumn{1}{l|}{92.65}  & \multicolumn{1}{l|}{92.72} & 92.87 & \multicolumn{1}{l|}{56.06}    & 61.454     & \multicolumn{1}{l|}{13.28} & 17.80 \\
Ours (K= 2)                 & \multicolumn{1}{l|}{98.90}  & \multicolumn{1}{l|}{91.43} & \multicolumn{1}{l|}{84.71} & \multicolumn{1}{l|}{92.35} & \multicolumn{1}{l|}{92.00}  & 91.88 & \multicolumn{1}{l|}{52.60}    & 58.48     & \multicolumn{1}{l|}{29.49} & 19.28 \\
OursV1 (K=1 \& 2)                 & \multicolumn{1}{l|}{99.23}  & \multicolumn{1}{l|}{92.96} & \multicolumn{1}{l|}{87.07} & \multicolumn{1}{l|}{92.93} & \multicolumn{1}{l|}{93.00}  & 93.04 & \multicolumn{1}{l|}{57.83}    & 63.31     & \multicolumn{1}{l|}{40.63} & 41.42 \\
OursV2 (K=1 \& 2)                 & \multicolumn{1}{l|}{99.05}  & \multicolumn{1}{l|}{92.20}  & \multicolumn{1}{l|}{85.92} & \multicolumn{1}{l|}{92.52} & \multicolumn{1}{l|}{92.17} & 92.37  & \multicolumn{1}{l|}{53.84}    & 59.33    & \multicolumn{1}{l|}{13.01} & 15.04
\\
OursV3 (K=1 \& 2)                 & \multicolumn{1}{l|}{99.12}  & \multicolumn{1}{l|}{92.69}  & \multicolumn{1}{l|}{85.95} & \multicolumn{1}{l|}{93.00} & \multicolumn{1}{l|}{92.08} & 92.47  & \multicolumn{1}{l|}{52.55}    & 58.07   & \multicolumn{1}{l|}{13.01} & 15.04
\\\hline
\end{tabular}
\end{center}
\vspace{-6mm}
\end{table*}
\begin{table}[]
\small
\begin{center}
\caption {An ablation study to investigate the impact of varying values of \(\alpha\) in Eq. \ref{weight} on the performance of our proposed method across different evaluation datasets. 1:1 verification average accuracy (Avg) for high-quality datasets, TAR@FAR=0.01\%  for IJB-B and Rank-1 accuracy for TinyFace are reported.  The backbone used here is Resnet18, respectively.}

\label{ablation_alpha}
\addtolength{\tabcolsep}{1pt}
\begin{tabular}{|l|l|l|l|l|}
\hline
Experiment & $\alpha$ & \multicolumn{1}{c|}{Avg} & \multicolumn{1}{c|}{TinyFace} & \multicolumn{1}{c|}{IJB-B} \\ \hline
1          & 0.05                  & 92.15                                                                                                                 & 54.45                                                                                       & 12.47                                                                                    \\
2 (Best)         & \textbf{0.1}                  &  \textbf{93.04}                                                                                                                &  \textbf{57.83}                                                                                       &  \textbf{40.63}                                                                                    \\
3          & 0.3                   & 91.71                                                                                                                 & 53.98                                                                                        & 15.35                                                                                    \\
4          & 0.5                   & 91.59                                                                                                                 & 53.86                                                                                        & 14.65  \\ \hline                                                                                 
\end{tabular}
\end{center}
\vspace{-8mm}
\end{table}


\begin{figure}
\begin{center}
\includegraphics[width=0.8\linewidth]{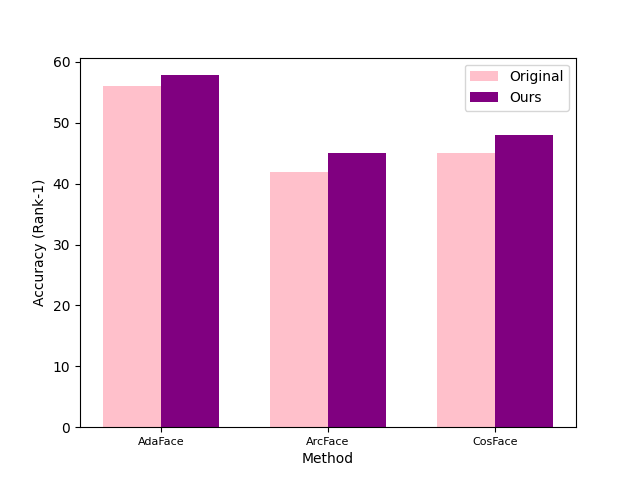}
\vspace{-4mm}
\end{center}
   \caption{The rank-1 face identification accuracy on the TinyFace dataset using the AdaFace \cite{kim2022adaface}, ArcFace \cite{deng2019arcface} and CosFace \cite{wang2018cosface} FR methods when there are just one model (K=1, original) and when our proposed method is applied (i.e., $K=1$ \& $K=2$, ours).}
\label{orthogonal}
\vspace{-4mm}
\end{figure}


\subsection{Performance Comparison}
Our proposed method's performance against SOTA studies has been assessed in Tables \ref{res18_highTiny}, \ref{res18IJB}, and \ref{high&tinyRes50}. 
According to the results, the gain on the high-quality dataset is less pronounced since these datasets mostly contain high-quality samples. Therefore, the current performance of other competitors is saturated. On the other hand, results show remarkable improvements on the more challenging benchmarks of IJB-B, IJB-C, and TinyFace. In the case of IJB-B and IJB-C, our method achieves over 10 percent improvement using the R18 backbone at $TAR@FAR=10^{-4}$. Also, over one percent improvement on TinyFace which shows that our method successfully maintains the performance on the high-quality samples and at the same time, it increases the discriminability among the hard instances.
It should be noted that the improvement is more sensible when we are using the weaker backbone, Resnet18, as shown in in Tables \ref{res18_highTiny} and \ref{res18IJB}.

\section{Ablation Study}
\subsection{Training with hard samples}
Solely training the $K^{th}$ model on the hard instances can lead to suboptimal solutions and overfitting.  Because discarding easy samples completely can be harmful, as they play a crucial role in relieving the representations from collapsing \cite{robbins2022effect}.
We investigated this effect by using only misclassified samples (from the training dataset) of the first model for training the second model ($K=2$). As it is shown in Table \ref{compare_ablation} (fourth row), the performance of our method degrades severely.
This reduction in discriminability can be attributed to 1) the reduction of the diversity of the data and 2) the extremely complex FR task when solely using hard instances.
\subsection{Discussion on Individual Model's Performance}
Boosting refers to combining different models to improve their overall performance. In this section, we evaluate the performance of individual models extracted from an ensemble model.
Our results, presented in Table \ref{compare_ablation}, indicate that the overall performance of the combined models is superior to that of any single model. This is because each model has its expertise in different groups of the training samples. Combining these models provides diverse discriminant information, resulting in a robust feature extractor with higher generalization. 

Furthermore, in the classical AdaBoost, each model is trained from scratch, which is unsuitable for CNN and might force the CNN to become overfitted on those samples with higher weight. Transferring the currently learned parameters to the next CNN helps the following CNN preserve the previous knowledge acquired during the learning process and reduces the computational cost. Table \ref{compare_ablation}, shows the comparison between our proposed method when the model corresponding to $K=2$ trained from scratch (OursV3) or fine-tuned from the previous model (OursV1). The higher performance of OursV1 demonstrates that transfer learning is significant in our approach. 

\subsection{Discussion on Re-weighting Samples During Training}
Fig. \ref{fig:norm_analysis} shows the easy and hard samples for two subjects from training dataset and their corresponding weights obtained by our method for $K=1$ and $K=2$. As it illustrates, the training of the first model indicates that frontal and high-quality faces are straightforward samples, resulting in lower corresponding weights compared to images with extreme poses or blurriness for training the second model. In the first model ($K=1$), all samples are assigned the same weight. However, for the second model ($K=2$), the weights of each sample are changed (based on Eq. \ref{prob} and \ref{weight}), enabling the model to prioritize more challenging and hard samples during the training of the second model.

\subsection{Orthogonality to Angular Criterion}
Our training framework includes ensemble learning when designing a FR module. We want to evaluate the effectiveness of this approach with various loss functions. Although our main experiments used the AdaFace loss function, our method represents an independent improvement on AdaFace. Specifically, we applied  two other SOTA loss functions, including ArcFace and CosFace (each with identical hyperparameters for margin and scale). Our results on the TinyFace dataset, as illustrated in Fig. \ref{orthogonal}, demonstrates that our approach enhances the feature embedding discriminability in all cases, indicating its independence from the choice of the training criterion.

\section{Conclusion}

To address the issue of imbalanced quality distribution in face recognition training datasets, we have proposed a novel approach that employs a sample-level weighting technique inspired by the traditional AdaBoost algorithm. By giving higher importance to the underrepresented tail samples during the training of a new model, our method is designed to improve the generalization performance of FR methods on such samples. The training loss function of an earlier model is used to update the sample training weights. If a sample is effectively trained by the first prior model, the weight associated with that sample is exponentially decreased, resulting in a negligible effect on the training of the next model and vice versa. The combination of different models, where each of them 
is an expert in different groups of training samples, leads to a robust classifier. Our approach successfully outperforms any SOTA FR single model in several challenging face benchmarks as depicted in the experimental section. We believe that our approach could be very helpful for large-scale unbalanced data training in each method.

\section{Acknowledgement}
This research is based upon work supported by the Office of the Director of National Intelligence (ODNI), Intelligence Advanced Research Projects Activity (IARPA), via IARPA R\&D Contract No. 2022-21102100001. The views and conclusions contained herein are those of the authors and should not be interpreted as necessarily representing the official policies or endorsements, either expressed or implied, of the ODNI, IARPA, or the U.S. Government. The U.S. Government is authorized to reproduce and distribute reprints for Governmental purposes notwithstanding any copyright annotation thereon.

{\small
\bibliographystyle{ieee}
\bibliography{egbib}
}

\end{document}